\documentclass[11pt,twocolumn,twoside]{IEEEtran}
\usepackage{amsmath}
\usepackage{url}
\PassOptionsToPackage{hyphens}{url}\usepackage{hyperref}
\usepackage[utf8]{inputenc}
\usepackage[margin=0.75in,headheight=0.45in]{geometry}
\usepackage[pdftex]{epsfig}
\usepackage{amsfonts}
\usepackage{amssymb}
\usepackage{fancyhdr}
\usepackage{nicefrac}
\include{graphicsx}

\begin{document}
\title{\vspace{0.2in}\sc Computing flood probabilities using Twitter: application to the Houston urban area during Harvey}
\author{Etienne Brangbour$^{1,2}$, Pierrick Bruneau$^{1}$\thanks{Corresponding author: P. Bruneau, pierrick.bruneau@list.lu $^1$Luxembourg Institute of Science and Technology \newline $^2$University of Geneva}, Stéphane Marchand-Maillet$^{2}$, Renaud Hostache$^1$, \newline Marco Chini$^1$, Patrick Matgen$^1$, Thomas Tamisier$^1$}

\maketitle
\thispagestyle{fancy}
\begin{abstract}
In this paper, we investigate the conversion of a Twitter corpus into geo-referenced raster cells holding the probability of the associated geographical areas of being flooded. We describe a baseline approach that combines a density ratio function, aggregation using a spatio-temporal Gaussian kernel function, and \emph{TFIDF} textual features. The features are transformed to probabilities using a logistic regression model. The described method is evaluated on a corpus collected after the floods that followed Hurricane Harvey in the Houston urban area in August-September 2017. The baseline reaches a F1 score of 68\%. We highlight research directions likely to improve these initial results.
\end{abstract}

\section{Introduction}

The seminal way of predicting whether a point in space and time will be flooded is to simulate the water flow and runoff resulting from the expected rainfall. For instance, water flow is simulated using a Digital Elevation Model in Lisflood-FP \cite{bates00}. More recently, an assimilation technique that allows to inject exogeneous observations, and adaptively update results of such simulations, was disclosed \cite{hostache15}. A 2D raster model of the region of interest is considered, with exogeneous probabilities of being flooded assigned to cells. The pixels in \emph{Synthetic Aperture Radar} (SAR) satellite images, classified using a hierarchical split-based approach \cite{chini_hierarchical_2017}, have been used as such proxy observations in \cite{hostache15}.

The usage of Twitter in the context of environmental hazards prevention and mitigation has been explored by several authors in the literature. For example, Twitter is used by Sakaki et al. \cite{sakaki_tweet_2013} to help damage detection and reporting in the context of Earthquake events. The TAGGS platform aims at using Twitter for flood impact assessment at a worldwide scale \cite{de_bruijn_taggs:_2017}. Twitter has also been used to monitor the spread of the seasonal flu disease \cite{chen_syndromic_2016}.

The objective of the present contribution is to perform and evaluate the conversion of a Twitter corpus into a map product analogous to that obtained from SAR images, for example. Data assimilation as described in \cite{hostache15} would then \emph{a priori} benefit from such multiple sources. Our specific contributions are the following: a Gaussian spatio-temporal kernel function that effectively drives feature vector construction, the combination of heterogeneous geographical information (Twitter fields and 2D geo-referenced raster data), and the application to a real-world use case. 


\section{Related Work}

Many papers about using Twitter for improving the response to natural disasters mainly focus on analyzing the spatial dimension of a Twitter corpus. Geographical information is present in tweets in the form of \emph{geotags} (i.e. discrete GPS coordinates) and \emph{bounding boxes} (surface rectangles). Among these, geotags are the most accurate \emph{a priori}, hence the most interesting. However, several studies report only about 1\% of all tweets holding a geotag \cite{middleton_real-time_2014}. This proportion has also been observed in a corpus we collected in relation to the Hurricane Harvey use case \cite{brangbour_extracting_2019}. 
In response to this lack of geographical information, several authors have focused on means to localize Twitter content. In \cite{de_bruijn_taggs:_2017}, the authors focus on toponym detection and disambiguation. This is of prime importance in their context, as their system collects content from any place in the world. 
Named Entity Recognition (NER) adapted to Twitter is used to extract geographical entities, then combined in a resolution index table. Schulz et al. \cite{schulz_multi-indicator_2013} combine geotags, NER results, bounding boxes, user information and emission time zones in a polygon stacking approach.
In our work, we focus on a smaller area of interest, that limits the need for toponym resolution. Also, we analyzed the geographical surfaces represented by the bounding boxes in a Twitter corpus collected for the Harvey use case, and concluded that approximately 17\% of the tweets in the corpus have geographical information relevant for a flood event \cite{brangbour_extracting_2019}. In the remainder of this paper, we focus on using this subset of the corpus as means to attach flood probabilities to cells in a 2D geographical raster model.
The most immediate way of detecting flood relatedness is to check for the presence of pre-defined keywords in tweet text \cite{middleton_real-time_2014}. For example, a corpus related to Hurricane Harvey has been collected by detecting the keywords \emph{Hurricane Harvey}, \emph{\#HurricaneHarvey}, \emph{\#Harvey} and \emph{\#Hurricane} during the event \cite{littman_hurricanes_2017}. In the context of a flu spread analysis, \cite{gao_mapping_2018} isolated a corpus matching a set of pre-defined keywords, and  manually annotated 6500 tweets in order to train a SVM classifier that further filters out false positives. Individual tweets are then aggregated w.r.t. space and time using a density rate function. \cite{lampos_nowcasting_2012} saves the effort of individual tweet annotation by averaging tweets at city and day scale w.r.t. their \emph{Term-Frequency-Inverse-Document-Frequency} (\emph{TFIDF}) representation \cite{joachims_text_1998}. This is a continuous version of the binary bag-of-word model used in \cite{gao_mapping_2018} for training a SVM classifier. A regression function linking observed rainfall to these feature vectors is then learned. In \cite{chen_syndromic_2016}, the authors explicitly account for the time dynamics by fitting a state-space model to the likelihood Twitter users have to catch flu. 


\section{Proposed Model}

Let us define the target variable $y$ as a binary flooding indicator. We consider a constant spatio-temporal resolution, with spatial and temporal coordinates $s$ and $t$, so that $ y_{s,t} = 1 \text{ if }(s,t) \text{ is flooded, 0 otherwise}$.
A given spatio-temporal cell is represented by its target variable $y_{s,t}$ and its feature vector $x_{s,t}$. These variables are linked according to the following logistic regression model:

\vspace{-4mm}
\begin{equation}
p(y_{s,t} = 1 | \phi_{s,t}) = \sigma(w^T \phi_{s,t}) \label{eq:logistic}
\end{equation}

with $\sigma(a) = (1 + \exp(-a))^{-1}$ and $\phi_{s,t} = (1,  x_{s,t}^T)^T$, therefore allowing for an intercept term in the adjustable weights vector $w$. For a given set of target values $\{ y_{s,t} \}$ and their associated feature vectors, there is a single optimal value $w^*$ to Eqn. \eqref{eq:logistic}, obtained by minimizing the following convex loss function \cite{lee_efficient_2006}:

\vspace{-4mm}
\begin{align}
\mathcal L(w) = c\| w \|_1 & - \sum_{s,t} {} y_{s,t} \ln \sigma(w^T \phi_{s,t}) \nonumber \\
 & + (1 - y_{s,t}) \ln (1 - \sigma(w^T \phi_{s,t})) \label{eq:loss}
\end{align}

with $c$ a positive regularization constant. The L1 penalty term ensures that the obtained solution is sparse, i.e. with as few non-zero coefficients in $w$ as possible.


\section{Feature Vectors} \label{sec:feature}

In order to compute Eqn. \eqref{eq:logistic} and minimize Eqn. \eqref{eq:loss}, feature vectors $x_{s,t}$ have to be defined. Let us consider a collection of tweets $\mathcal C_N$:

\vspace{-4mm}
\begin{equation}
\mathcal C_N = \biggl \{ \{s_n, d_n, t_n, \Omega_n \}_{n \in 1 \dots N} \biggr \} \label{eq:tweetdef}
\end{equation}

where $s_n$ is the spatial index of the $n^\text{th}$ tweet, $d_n$ its spatial dispersion (homogeneous to a standard deviation), $t_n$ the temporal index of the tweet, and $\Omega_n$ the array of phrases in the tweet. The geographical information in tweets is composed of discrete geotags, as well as bounding boxes. The latter are abstracted by the dispersion variable in this section. The computation of this dispersion out of actual data is presented in Section \ref{sec:experiments}.
In Gao et al., the authors define the scalar \emph{Social Media Event Rate (SMER)} of a spatio-temporal cell $x^\textit{SMER}_{s,t}$ as \cite{gao_mapping_2018}:

\vspace{-2mm}
\begin{equation}
    x^\textit{SMER}_{s,t} = \frac { \sum_{n=1}^N {K(s, s_n, d_n) I(t, t_n) z_n}}{ \sum_{n=1}^N {K(s, s_n, d_n) I(t, t_n)}} \label{eq:smer}
\end{equation}

with $K$ and $I$ the spatial and temporal kernel functions, respectively. The random variable $z_n$ equals 1 if $\Omega_n$ overlaps a query defined by the user, 0 else. In \cite{gao_mapping_2018}, the authors use the Epanechnikov kernel function to model the spatial coordinates. For an explicit account of the dispersion information, we rather use a Gaussian spatial kernel function:

\vspace{-4mm}
\begin{equation}
    K(s, s_n, d_n) = (2 \pi d_n^2)^{-\frac 1 2} \exp \biggl ( - \frac{\| s - s_n \|_2^2} {2 d_n^2} \biggr ).
\end{equation}

The spatial L2 norm is computed by combining differences in latitude and longitude as independent dimensions.
An indicator function is used as the temporal kernel: $I(t, t_n) = 1 \text{ if } t = t_n, 0 \text{ else}$.
Alternatively, let us consider the union of all $V$ phrases present in $\mathcal C_N$. Assuming an arbitrary order of these $V$ tokens defines a $V$-dimensional feature space. The \emph{term frequency} $\text{tf}_v(n)$ is simply the number of times token $v$ appears in $\Omega_n$. The \emph{term frequency} and \emph{inverse document frequency} of phrase $v$ are, respectively:

\vspace{-4mm}
\begin{align}
 & \text{tf}_v(n) = \bigl | \{ o = v | o \in \Omega_n \} \bigr | \nonumber \\
 & \text{idf}_v = \ln (1 + N) - \ln(1 + N_v) + 1 \nonumber
\end{align}

with $N_v = \bigl | \{ n \in 1 \dots N | \text{tf}_v(n) > 0 \} \bigr |$ the document frequency of phrase $v$ in $\mathcal C_N$. The \emph{TFIDF} feature vector for tweet $n$ is then defined as $\rho_n = ( \text{tf}_v(n) \times \text{idf}_v )_{v \in V}$ \cite{joachims_text_1998}. As means to build a feature vector that characterizes a spatio-temporal raster cell, we adapt Eqn. \eqref{eq:smer} as follows:

\vspace{-2mm}
\begin{equation}
    x^\textit{TFIDF}_{s,t} = \frac { \sum_{n=1}^N {K(s, s_n, d_n) I(t, t_n) \rho_n}}{ \sum_{n=1}^N {K(s, s_n, d_n) I(t, t_n)}} \label{eq:tfidf}
\end{equation}

In practice, we replace the scalar query overlap variable $z_n$ in Eqn. \eqref{eq:smer} with V-dimensional $\rho_n$. Performing this spatio-temporal aggregation is also a way to mitigate the fact that \emph{TFIDF} vectors are very sparse in the case of tweets. $x^\textit{TFIDF}_{s,t}$ instead yields denser representations, that are more easily handled by classification models in general - and Eqn. \eqref{eq:logistic} in particular.
Both $x^\textit{SMER}_{s,t}$ and $x^\textit{TFIDF}_{s,t}$ are tested in Section \ref{sec:experiments}.


\section{Experiments} \label{sec:experiments}

\subsection{Methodology} \label{sec:method}

\begin{figure}
\begin{center}
\epsfxsize=0.75\hsize \epsfbox{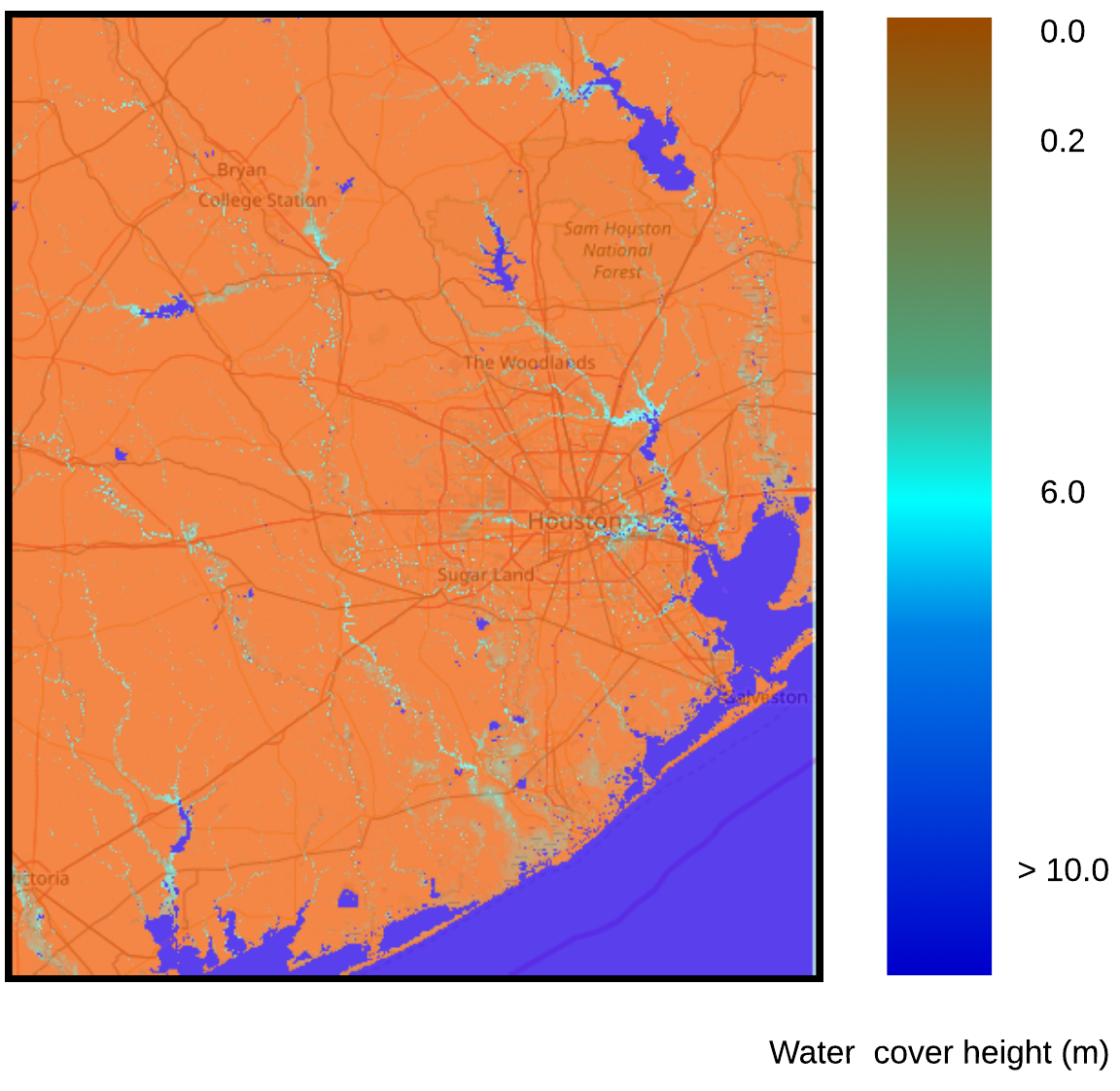}
\end{center}
\caption{Flooding map for the Houston urban area on 2017/08/30.}
\label{fig:flood}
\end{figure}

Hurricane Harvey has affected the Houston urban region between mid-August and mid-September 2017, with a peak around the 30${}^\text{th}$ of August. A corpus of tweets collected for the Harvey event has been made available shortly after \cite{littman_hurricanes_2017}. It features tweets matching any of the phrases \emph{Hurricane Harvey}, \emph{\#HurricaneHarvey}, \emph{\#Harvey} and \emph{\#Hurricane}. Alternatively, using the Twitter APIs, we collected our own corpus of 7.5M tweets. In order to match the scope and objectives disclosed in the introduction, especially regarding content localization, we did not use textual query filters, and collected all content matching the spatial bounds of the Houston urban surroundings, and the temporal bounds mentioned above. The spatial area of interest has been determined according to prior analyses of Hurricane Harvey impacts \cite{matgen_integrating_2017}. Details about the corpus collection, pre-processing, and descriptive analysis are described at length in \cite{brangbour_extracting_2019}. 
Figure \ref{fig:flood} shows a flooding map of the Houston urban area for the 30${}^\text{th}$ of August. It has been obtained by running the Lisflood-FP model \cite{bates00}. The raster matrix displayed has been scaled to dimensions $1225 \times 1450$ pixels (i.e. approximately 1.78M pixels in total), with resolution ${2.10^{-3 }}^{\circ}$ (approximately 240m) per pixel.

In this matrix, permanent water is indicated by a large value (i.e. 999). Hence we exclude permanent water pixels from our analysis by considering only pixels with water height less than 10m. This leaves a database of 1.47M pixels. Then, following \cite{bates00}, only pixels associated with water height greater than 0.2m are considered as flooded. This defines 80\% non-flooded and 20\% flooded pixels. The spatio-temporal binary target variable used in Eqn. \eqref{eq:logistic} takes its ground truth values from these pixels. 
For our experiments, we consider daily temporal resolution, i.e. $I(t, t') = 1 \text{ iif day}(t) = \text{day}(t')$ in Eqns. \eqref{eq:smer} and \eqref{eq:tfidf}.

Our corpus features 319280 tweets matching the 30${}^\text{th}$ of August. Among those, we further select 55214 tweets which hold geographical information sufficiently accurate to be retained in the analysis (see \cite{brangbour_extracting_2019} for details about the selection method). We consider this subset as $\mathcal C_N$ (see Eqn. \eqref{eq:tweetdef}).
To build $x^\textit{SMER}_{s,t}$ (Eqn. \eqref{eq:smer}) using this collection, we test the two following query filters: the keywords used in \cite{littman_hurricanes_2017}, and the keywords \emph{flood} and \emph{harvey}, that intuitively made sense to us.
Rectangular bounding boxes are attached to tweets. The dispersion $d_n$ in Eqn. \eqref{eq:tweetdef} is computed as the square root of the product of the bounding box half-width and half-height. We smooth these values by setting $d_s = \max(d_s, 10^{-3})$ (in degrees, i.e. approximately 100m). 
Preprocessing is recommended prior to building \emph{TFIDF} vectors \cite{lampos_nowcasting_2012}. First, URLs and user names are removed from the tweets. Then camel case words, commonly encountered on Twitter, are split as single words (e.g. \emph{HurricaneHarvey} becomes \emph{Hurricane Harvey}). Then punctuation and stop words are removed. Emoticons are not removed, and considered as words. Finally, we apply Porter stemming \cite{porter_algorithm_1980} to all words, that effectively normalizes the text (e.g. \emph{flood}, \emph{flooding} and \emph{flooded} all become \emph{flood}). Obtained feature vectors are normalized so that $\| \rho_n \|_2 = 1, \forall n$ (see Section \ref{sec:feature} for the definition of $\rho_n$).

Following the recommendations in \cite{lampos_nowcasting_2012}, we consider the set of 1 and 2-grams as the $V$-dimensional \emph{TFIDF} space. Simply put, dimensions in $x^\textit{TFIDF}_{s,t}$ are associated to single words in the 1-gram model, whereas 2-word phrases are also included in the 2-gram model.
This yields a very high-dimensional space (29K with 1-grams alone, 203K with 1 and 2-grams). Actually many of these tokens are present only a few times in $\mathcal C_N$: as suggested in \cite{lampos_nowcasting_2012}, we retain features $v$ with $N_v > 10$. This yields 3327 1-grams, and 1454 2-grams.
The regularization hyper-parameter $c$ in Eqn. \eqref{eq:loss} is optimized using 5-fold cross-validation. The loss is optimized using the SAGA solver \cite{defazio_saga:_2014}. For each feature vector type, the average F1 score shown in Table \ref{tab:f1} has been computed according to 20 independent runs. For a given run, we have drawn balanced samples of 20K pixels as the training set in Eqn. \eqref{eq:logistic}, as well as balanced samples of 2K pixels as the test sets. Then feature vectors are built according to Eqn. \eqref{eq:smer} and \eqref{eq:tfidf}. In particular, $x^\textit{TFIDF}_{s,t}$ is renormalized to unity after aggregation.

\vspace{-3mm}
\subsection{Results}

\begin{table}
    \begin{center}
    \begin{tabular}{|l|l|}
      \hline
      Feature Vector & F1 score \\
      \hline
      \emph{SMER} &  \\
      Hurricane Harvey, \#HurricaneHarvey, \dots \cite{littman_hurricanes_2017} & 0.58 $\pm$ 0.01 \\
      flood, harvey & 0.59 $\pm$ 0.01 \\
      \hline
      \emph{TFIDF} & 0.68 $\pm$ 0.01 \\
      \hline
    \end{tabular}
    \end{center}
    \caption{Test F1 errors obtained using $x^\textit{SMER}$ and $x^\textit{TFIDF}$.}
    \label{tab:f1}
\end{table}

For our experimental setup (ie. 2 balanced classes for training and testing), the F1 score expected by chance is 0.5. In Table \ref{tab:f1}, we see that \emph{SMER} features yield moderate improvement over chance. We get better results using our intuitive keyword set \emph{flood, harvey}, but the difference is not statistically significant. On the other hand, the \emph{TFIDF} feature vector brings approximately 10\% performance boost, well beyond significance levels.
As we mentioned in Section \ref{sec:method}, the \emph{TFIDF} vector has very high dimensionality, so we focus on identifying and inspecting its most relevant features for the classification problem at hand. First, as we used L1 penalty in Eqn. \eqref{eq:loss}, we obtain a drastic reduction of the number of features used in the model. In the end, 896 features have non-zero weight in average, reduced to 247 if we take the median of our experiments. This is approximately 5\% of the total number of features fed to the model. 

The magnitude of weights in $w$ can act as simple relevance scores of the respective features \cite{lecun_optimal_1990}. As means to aggregate our independent experimental runs (20), we first rank the features in decreasing relevance order. Then we normalize the scores in $[0,1]$ using $\nicefrac {(\kappa - \text{rank})} \kappa$, with $\kappa$ the number of features selected in the run. We aggregate runs by averaging these normalized scores. 
If we consider only features present in all selected feature sets, we get the following features, ordered by decreasing score: \emph{sad, fake, via, eye, today, oop, drive, dalla}. If we extend to the union of all features, the best 20 are: \emph{sad, fake, basic, via, eye, today, told, true, flow, garbag, old, ah, flood, realli, peopl, hockley, learn, guadalup rv, final, work}.
Manually inspecting the database, an example of tweet containing the word \emph{sad} is \emph{"mann, pray for my city houston, it's so sad seeing houston like dis!"} \emph{sad} is then possibly a positive (though unexpected \emph{a priori}) marker of flood relatedness. On the other hand, the token \emph{fake} seems mostly related to fake news discussions, that could possibly be negatively correlated with the flood concept, i.e. people affected by the flood would have other concerns than talking about fake news. Some highly relevant tokens such as \emph{flood} or \emph{flow} make sense intuitively, where the role of others is hard to figure out (e.g. \emph{eye} or \emph{old}). Finally, it is worth noting that only one 2-gram appears in our highlights. This is quite natural, as 2-grams are much more rare than 1-grams on average, and have thus much less leverage on model fitting \emph{a priori}. More striking is the absence of the token \emph{harvey}. One possible explanation is that it used as much in general posts about the hurricane, as by people affected by floods. Its discriminative power would then be lessened.

\section{Conclusion}

In this paper, we established a baseline for the performance in using Twitter posts to estimate whether a cell in a 2D raster grid is flooded. This baseline obtained a F1 score of 0.68, which is already very significantly better than chance, but also leaves much room for improvement.
For this baseline, we use \emph{TFIDF} features, which have recently been superseded by dense representations obtained from deep neural networks. For example, the latent vectors of a BiLSTM model trained for hashtag prediction have been used in the context of Twitter text \cite{dhingra_tweet2vec:_2016}. However, averaging is not necessarily meaningful in such spaces, so simple stacking strategies such as presented in this paper may not apply directly.
For future work, we will also consider the benefit from fields other than text and geographical information, such as the presence of an attached image \cite{peters_investigating_2015}, its content \cite{bischke_multimedia_2017}, or extracting geographical cues from text \cite{middleton_real-time_2014, de_bruijn_taggs:_2017}.

\section{Acknowledgements}

This work was performed in the context of the Publimape project, funded by the CORE programme of the Luxembourgish National Research Fund (FNR).

\bibliographystyle{ieeetr}
\bibliography{ci_references}

\end{document}